%% file: main.tex
\documentclass[10pt,twocolumn,letterpaper]{article}

\usepackage[pagenumbers]{cvpr} 

\definecolor{cvprblue}{rgb}{0.21,0.49,0.74}
\usepackage[pagebackref,breaklinks,colorlinks,allcolors=cvprblue]{hyperref}
\usepackage{adjustbox}
 \usepackage{multirow}
\usepackage{amsmath}
 \usepackage{tcolorbox}
 \usepackage{etoolbox}
 \usepackage{float}
 \usepackage{xcolor}
\usepackage{blindtext}
\tcbuselibrary{listingsutf8}
 \usepackage[numbers,sort&compress]{natbib}
 \newcommand{\smallcircled}[1]{\textcircled{\scriptsize#1}}
\def\paperID{22} 
\def\confName{CVPR}
\def\confYear{2025}
\usepackage{fancyhdr}

\title{Agglomerating Large Vision Encoders via Distillation for VFSS Segmentation}

\author{
Chengxi Zeng,
Yuxuan Jiang,
Fan Zhang,
Alberto Gambaruto,
Tilo Burghardt
\\
{\tt\small \{simon.zeng, yuxuan.jiang, Fan.Zhang, alberto.gambaruto, tb2935\}@bristol.ac.uk}
\\
\textit{University of Bristol, UK}
}

\begin{document}
\maketitle
\begin{abstract}
    The deployment of foundation models for medical imaging has demonstrated considerable success. However, their training overheads associated with downstream tasks remain substantial due to the size of the image encoders employed, and the inference complexity is also significantly high. Although lightweight variants have been obtained for these foundation models, their performance is constrained by their limited model capacity and suboptimal training strategies. In order to achieve an improved tradeoff between complexity and performance, we propose a new framework to improve the performance of low complexity models via knowledge distillation from multiple large medical foundation models (e.g., MedSAM, RAD-DINO, MedCLIP), each specializing in different vision tasks, with the goal to effectively bridge the performance gap for medical image segmentation tasks. The agglomerated model demonstrates superior generalization across 12 segmentation tasks, whereas specialized models require explicit training for each task. Our approach achieved an average performance gain of 2\% in Dice coefficient compared to simple distillation.
\end{abstract} 

\section{Introduction}
\label{sec:intro}

Medical images differ from natural images in several fundamental aspects. They often display lower contrast and exhibit high content complexity across various modalities. The interpretation of medical images primarily focuses on anatomical structures, tissues, and specific pathologies rather than encompassing a diverse range of objects. Medical image segmentation is a complex downstream task that demands a comprehensive understanding of visual modalities and the capability for anatomical localization. This is often achieved by specialist medical image models with explicit training or fine-tuning on a single segmentation task. More recently, large foundation models, including Segment Anything (SAM)~\cite{kirillov2023segany, ravi2024sam2}, have achieved significant success in general image segmentation tasks. Their remarkable zero-shot capabilities have enabled a wide array of downstream applications, including medical image segmentation~\cite{Ma2023SegmentAI, Ma2024SegmentAI}, among others. However, it is noted that these large models face challenges in real-time inference, primarily due to their bulky encoders.

Recent research~\cite{Zhou2023EdgeSAMPD, Wang2023RepViTSAMTR, Xiong2023EfficientSAMLM} aimed at enhancing the efficiency of SAM has primarily employed Teacher-Student knowledge distillation techniques~\cite{Hinton2015DistillingTK}. These approaches develop lightweight image encoders trained on reduced subsets of the original dataset. This strategy is also applicable in the context of medical imaging~\cite{Ma2023SegmentAI, ma2024efficientmedsamssegmentmedical}.

Conventional knowledge distillation methods have demonstrated efficacy in reducing model size while preserving acceptable performance metrics. However, their effectiveness often diminishes as the size of the student models diverges. This discrepancy manifests as a notable performance gap between the teacher and student models, which can be attributed to the inherent limitations of simple embedding or label matching strategies.




\begin{figure}[t]
\captionsetup{belowskip=-10pt}
\centering
\includegraphics[width=230pt]{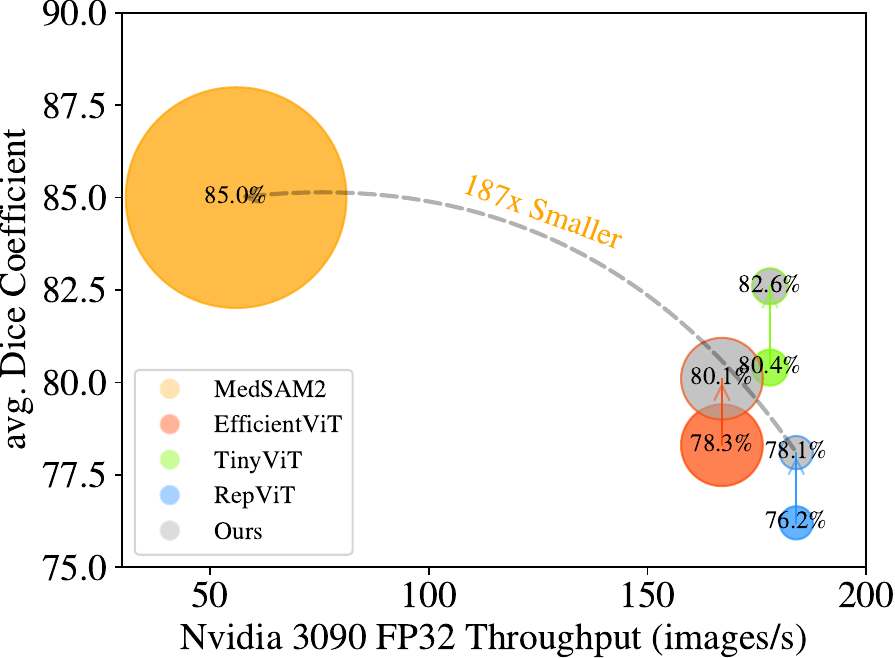}
\caption{
Efficient models have shown performance discrepancies compared to fine-tuned foundation model (MedSAM) during distillation; our approach enhances performance without introducing extra parameters. The size of the circles represents the model size.}
\label{fig:data efficiency}
\end{figure}

To address this limitation, inspired by multi-teacher distillation methods developed for robotic vision~\cite{Shang2024TheiaDD} and low-level~\cite{Jiang2024MTKDMK} vision applications, we propose a multi-model agglomeration-based knowledge distillation framework that leverages knowledge from multiple expert models, each specializing in distinct aspects of the segmentation task, to enrich the distilled representations (i.e., MedCLIP~\cite{Wang2022MedCLIPCL} for modality understanding, RAD-DINO~\cite{Heinrich2024RADIOv25IB} for localization, and MedSAM~\cite{Ma2024SegmentAI} for segmentation). By aggregating diverse expertise, the approach effectively captures a wider range of visual features, contextual nuances, and spatial relationships inherent in medical images. This results in a higher-performing and more robust distilled lightweight model, improving generalization across varied segmentation tasks. Our contributions are summarized as follows:
\begin{itemize} 
\item \textbf{An effective multi-model agglomeration framework} that leverages the strengths of advanced medical vision models, enhancing the distilled model's capabilities in medical image segmentation without introducing additional training parameters. As far as we know, this is the first time that this type of approach has been applied in the context of medical image segmentation.

\item \textbf{A new training strategy} that balances losses between multiple teachers, along with standardization techniques that incorporate teacher models more effectively during training. 
\item \textbf{Lighter encoders} are up to $187\times$ smaller but improve average segmentation accuracy by 2\% compared to simple distilled encoders, as demonstrated in~\autoref{fig:data efficiency}.
\end{itemize}

\section{Related Work}
\label{sec:related work}
\noindent\textbf{Multi-Model Agglomeration.}
Agglomerative models focus on consolidating knowledge from various pretrained models into a unified vision encoder. Preliminarily, SAM-CLIP~\cite{Wang2023SAMCLIPMV} combines the SAM model for segmentation with CLIP; however, the lack of robust dense models like DINO~\cite{Oquab2023DINOv2LR} has led to diminished accuracy in dense downstream tasks such as segmentation. Noteworthy recent approaches include AM-RADIO~\cite{Ranzinger2023AMRADIOAV, Heinrich2024RADIOv25IB}, which implemented label-free distillation from multiple teacher models but demonstrated performance that varied with resolution across different tasks. Theia~\cite{Shang2024TheiaDD} introduces a multi-teacher distillation approach tailored for robotic vision tasks. Another line of work, Eagle~\cite{Shi2024EagleET}, employs a mixture of encoders to improve multimodal LLMs.

\vspace{5pt}

\noindent\textbf{Large Foundation Models.}
Vision Transformers (ViT) have shown exceptional performance across a range of computer vision tasks. Following this, models like DINOv2~\cite{Oquab2023DINOv2LR} and CLIP~\cite{Radford2021LearningTV} have made significant contributions to self-supervised learning and vision-language integration. In the medical domain, RAD-DINO~\cite{PerezGarcia2024RADDINOES} and MedCLIP~\cite{Wang2022MedCLIPCL} have gained notable traction, demonstrating effectiveness in areas such as histopathology image analysis~\cite{song2024morphological}, ophthalmology~\cite{ANTAKI2023100324}, and X-ray imaging~\cite{Abdulaal2024AnXI}. While these pretrained models are recognized for their strong zero-shot generalization abilities, their substantial encoder sizes present challenges for applications requiring real-time processing.

\vspace{5pt}

\noindent\textbf{Efficient Segment Anything Models (SAM).} 
Various strategies have been developed to enhance the efficiency of SAM models, including FastSAM~\cite{Zhao2023FastSA}, MobileSAM~\cite{Zhang2023MobileSAMv2FS}, EdgeSAM~\cite{Zhou2023EdgeSAMPD}, RepViTSAM~\cite{Wang2023RepViTSAMTR}, and EfficientSAM~\cite{Xiong2023EfficientSAMLM}. These models utilize a range of techniques, such as knowledge distillation and lightweight architectures, to minimize model size while preserving performance. Similarly, LiteMedSAM~\cite{Ma2023SegmentAI} and EfficientMedSAM~\cite{Ma2024SegmentAI} focus on enhancing efficiency in Medical SAMs.

\section{Methodology}
\label{sec: Methodology}
\begin{figure*}[ht]
    \captionsetup{belowskip=-10pt}
    \centering
    \includegraphics[width=450pt]{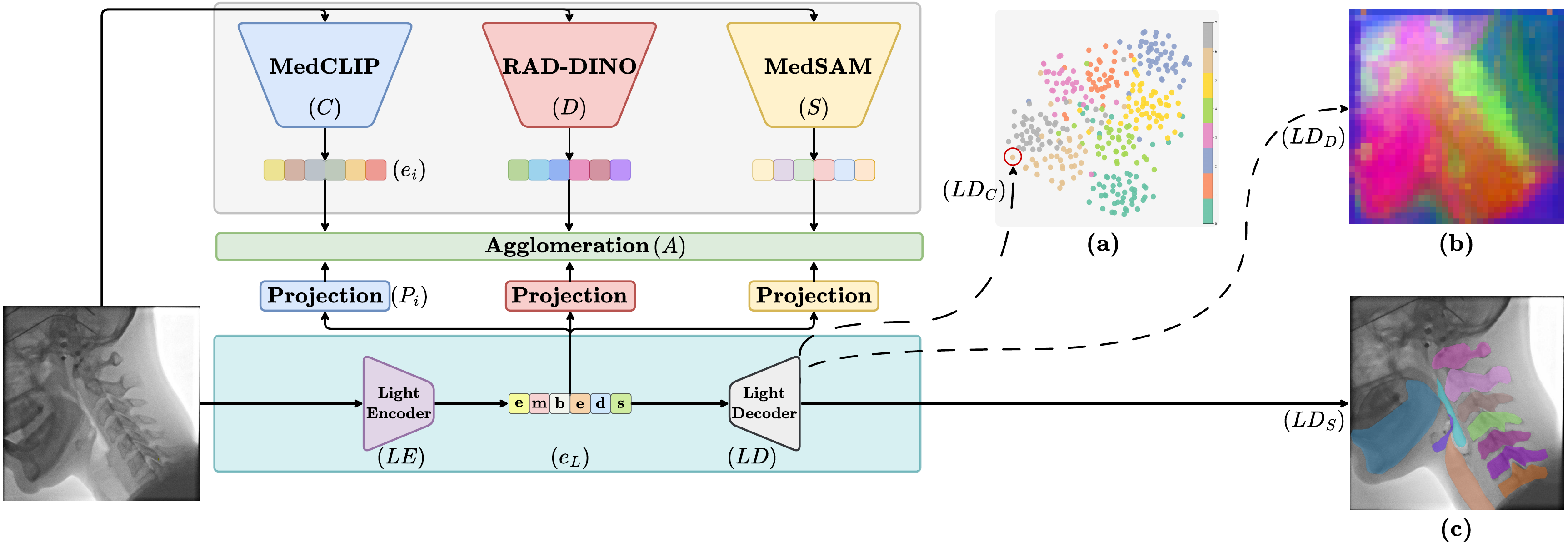}
    \caption{
The teacher models' features are agglomerated into student features. The learned representations can be decoded by multiple Light Decoders ($LD$) for general tasks: (a) Clustering visualization, (b) PCA visualization, and (c) SAM-decoded Instance Segmentation.}
 
\label{fig:methodology}
\end{figure*}
\autoref{fig:methodology} illustrates the proposed framework, which adheres to the standard encoder-decoder (E-D) architecture of SAM. This architecture facilitates a versatile design for encoder agglomeration within the encoded latent space, enabling the integration of diverse feature representations. The encoder is distilled from a variety of teacher vision models, thereby encapsulating a rich array of representations. The decoder preserves the original design of SAM, adeptly processing both image and prompt embeddings.

Theoretically, any encoder-decoder (E-D) architecture can be integrated into this framework. To enhance the efficiency of our model, we utilize lightweight encoder models, denoted as $LE$. The segmentation decoder ($LD_S$) is inherited from SAM2, as it is already very lightweight.

\subsection{Multi-Model Agglomeration via Distillation}

Knowledge agglomeration focuses on extracting unique knowledge from various foundational teacher models ($T$) to create a cohesive student model. 

Formally, let $x$ represent an input image. The teacher model generates feature representations as:
\begin{equation}
    \left[e_{1}, e_{2}, \dots, e_{i}\right] = \left[T_1(x), T_2(x), \dots, T_i(x) \right],
\end{equation}
where $e_i$ are distinct teacher embeddings whose spatial patch tokens correspond to the individually projected embeddings from the student encoder. This approach differs from AM-RADIO~\cite{PerezGarcia2024RADDINOES}, which utilizes the \texttt{CLS} token $\in \mathbb{R}^{1 \times N}$. In this work, we adhere to Theia~\cite{Shang2024TheiaDD} and exclusively employ the \texttt{Patch} token $\in \mathbb{R}^{(d-1) \times N}$ for simplicity, as SAM does not incorporate \texttt{CLS} tokens.

Each projection head from the student embedding ($e_L$) is a multi-layer perceptron (MLP), defined as:
\begin{equation}
    z_i = P_i(e_L) = \sigma(W^{T} e_L + b^{T}),
\end{equation}
where $\sigma$ is a non-linear activation function, and $W^{T}$, $b^{T}$ are learnable parameters specific to teacher $T$.

The distillation loss $\mathcal{L}$ for aligning the student and teacher representations is computed by comparing projected student features with teacher features, using an appropriate similarity or distance metric (e.g., mean squared error or cosine similarity loss):
\begin{equation}
\mathcal{L}^{LE} = \sum_{i=1}^{T} \ell_i(z_i^{P}, e_i^{T}),
\end{equation}
where $\ell_i$ is the loss between teacher and student features.

\subsection{Agglomeration Strategies}
\label{sec: Agglomeration Strategies}
When training a student model with multiple teacher encoders, balancing the contribution from each teacher's loss function effectively is crucial. We propose two approaches: (1) MLP-based Loss Balancing and (2) Attention-based Loss Balancing to be the agglomeration layer.

MLP-based Loss Balancing weights multiple teacher losses adaptively during the distillation process. The adaptive weighting is formed as:
\begin{equation}
\alpha_i^{MLP} = \text{softmax}\left(W_2 \sigma(W_1 \mathcal{L} + b_1) + b_2\right), 
\end{equation}
where $W_1 \in \mathbb{R}^{h \times T}$, $b_1 \in \mathbb{R}^{h}$, $W_2 \in \mathbb{R}^{1 \times h}$, and $b_2 \in \mathbb{R}$ are learnable parameters, and $h$ is an arbitrary hidden size.


For Attention-based Loss Balancing, we introduce a learnable query vector $q \in \mathbb{R}^{d}$ from the student embedding and key vectors $k_i \in \mathbb{R}^{d}$, each corresponding to the teacher encoder $T$. Attention weights are computed using scaled dot-product attention:
\begin{equation}
\alpha_i^{Attn} = \text{softmax}\left(\frac{q^\top k_i}{\sqrt{d}}\right),
\end{equation}
This method adaptively quantifies each encoder’s contribution based on its similarity to the query vector.

The final loss in either case, integrating all teacher losses with MLP- or attention-derived weights, is:
\begin{equation}
\mathcal{L} = \sum_{i=1}^{T}\alpha_i \ell_i.
\end{equation}



\subsection{Implementation Details} 

\noindent\textbf{Models.} In this work, we specifically utilize three pretrained (large) encoders as teachers from MedCLIP ($C$)~\cite{Wang2022MedCLIPCL} due to their generalizability in medical image understanding, RAD-DINO ($D$)~\cite{Heinrich2024RADIOv25IB} for its outstanding performance in localization, and MedSAM2 ($S$)~\cite{Ma2024SegmentAI} for its effectiveness in downstream segmentation. The student candidates evaluated in this experiment include three SOTA architectures, TinyViT~\cite{tiny_vit}, RepViT~\cite{Wang2023RepViTSAMTR} and EfficientViT~\cite{cai2022efficientvit}

\vspace{5pt}
\noindent\textbf{Dataset.} We use a challenging medical segmentation dataset, VFSS-5K~\cite{Zeng2025TuningVF}, which consists of comprehensive annotations of approximately 12 anatomical structures per frame in the videofluoroscopy video data, including the bolus, pharynx, trachea, epiglottis, mandible, and cervical vertebrae (C1-C7), totaling around 5,000 instance annotations. The training and test sets are split in a 7:3 ratio.

\vspace{5pt}
\noindent\textbf{Metrics.} reported in this paper are the common Dice coefficient~\cite{Dice1945MeasuresOT} for segmentation area accuracy and HD95~\cite{Goldlcke2014VariationalA} for surface distance accuracy.

\vspace{5pt}
\noindent\textbf{Training details.} The Multi-Model Distillation process consists of 100 epochs dedicated to training the student encoder, followed by an additional 100 epochs for aligning the decoder with the segmentation ground truth. We employ the AdamW optimizer with parameters $\beta_1 = 0.9$, $\beta_2 = 0.999$, and a weight decay of 0.1. The learning rate is set to decay from $1 \times 10^{-4}$ to $1 \times 10^{-5}$.

\subsection{Results and Discussion}

\begin{tcolorbox}[colframe=black, colback=gray!5, boxrule=0.5mm, arc=4pt, auto outer arc]
\textit{\textbf{Q1.}} Distilled Efficient foundation models \textit{V.S.} Specialist Models?
\end{tcolorbox}

\begin{table}[ht]
\caption{Comparing the efficiency of the models; '$-m$' denotes our multi-encoder distilled model. $\colorbox{blue!50}{DarkBlue}$ indicates best specialist model and $\colorbox{blue!10}{LightBlue}$ indicates best efficient models.}
\label{tab: student models}
\begin{adjustbox}{width=240pt,center}
\begin{tabular}{@{}ccccc@{}}
\toprule

\multicolumn{1}{c|}{Models} &  \multicolumn{1}{c|}{Params}& \multicolumn{1}{c|}{FLOPs} & \multicolumn{1}{c|}{avg. Dice $\uparrow$} & \multicolumn{1}{c}{avg. HD95 $\downarrow$} \\ \midrule

\multicolumn{1}{l|}{TransUNet~\cite{Chen2021TransUNetTM}}    & \multicolumn{1}{c|}{105.3M} & \multicolumn{1}{c|}{29.3G} & \multicolumn{1}{c|}{0.859} & \multicolumn{1}{c}{7.451}     \\
\multicolumn{1}{l|}{Video-TransUNet~\cite{Zeng2022VideoTransUNetTB}}    & \multicolumn{1}{c|}{110.5M} & \multicolumn{1}{c|}{40.4G} & \multicolumn{1}{c|}{0.880} & \multicolumn{1}{c}{6.916}     \\
\multicolumn{1}{l|}{SwinUNet~\cite{Cao2021SwinUnetUP}}    & \multicolumn{1}{c|}{27.1M} & \multicolumn{1}{c|}{16.1G} & \multicolumn{1}{c|}{0.848} & \multicolumn{1}{c}{10.290}     \\
\multicolumn{1}{l|}{Video-SwinUNet~\cite{zeng2023video}}    & \multicolumn{1}{c|}{48.9M} & \multicolumn{1}{c|}{25.8G} & \multicolumn{1}{c|}{\colorbox{blue!50}{0.899}} & \multicolumn{1}{c}{\colorbox{blue!50}{6.237}}     \\
\midrule
\multicolumn{1}{l|}{MedSAM2~\cite{Ma2024SegmentAI}}    & \multicolumn{1}{c|}{224.4M} & \multicolumn{1}{c|}{60.1G} & \multicolumn{1}{c|}{0.849} & \multicolumn{1}{c}{8.974}     \\
\midrule
\multicolumn{1}{l|}{TinyViT}      & \multicolumn{1}{c|}{6.0M}& \multicolumn{1}{c|}{4.4G} & \multicolumn{1}{c|}{0.804} & \multicolumn{1}{c}{14.436}   \\
\multicolumn{1}{l|}{TinyViT-m}      & \multicolumn{1}{c|}{6.0M}& \multicolumn{1}{c|}{4.4G} & \multicolumn{1}{c|}{\colorbox{blue!10}{0.826}} & \multicolumn{1}{c}{\colorbox{blue!10}{12.641}}   \\
\multicolumn{1}{l|}{RepViT}  & \multicolumn{1}{c|}{5.1M}& \multicolumn{1}{c|}{4.1G} & \multicolumn{1}{c|}{0.762} & \multicolumn{1}{c}{16.434}  \\ 
\multicolumn{1}{l|}{RepViT-m}  & \multicolumn{1}{c|}{5.1M}& \multicolumn{1}{c|}{4.1G} & \multicolumn{1}{c|}{\colorbox{blue!10}{0.781}} & \multicolumn{1}{c}{\colorbox{blue!10}{15.123}}  \\ 
\multicolumn{1}{l|}{EfficientViT}    & \multicolumn{1}{c|}{30.8M}& \multicolumn{1}{c|}{8.9G} & \multicolumn{1}{c|}{0.783} & \multicolumn{1}{c}{15.099}  \\
\multicolumn{1}{l|}{EfficientViT-m}    & \multicolumn{1}{c|}{30.8M}& \multicolumn{1}{c|}{8.9G} & \multicolumn{1}{c|}{\colorbox{blue!10}{0.801}} & \multicolumn{1}{c}{\colorbox{blue!10}{14.871}}  \\
\bottomrule
\end{tabular}
\end{adjustbox}
\vskip -10pt
\end{table}

We first compare the resulting specialist models based on our distillation framework with those obtained from simple knowledge distillation \cite{Hinton2015DistillingTK}, alongside the original large foundation model, MedSAM2. The results in \autoref{tab: student models} show the impressive generalizability of our models, which achieve an average 2\% improvement in the Dice coefficient for all three efficient encoders. The complexity-performance trade-offs of all these models are also illustrated in \autoref{fig:data efficiency}. These results highlight the effectiveness of our approach and the potential of efficient models to achieve competitive performance across various tasks. Although the specialist model shows overall better results, it is trained on each segmentation task, and its inference overhead is much higher than that of efficient models.


\begin{tcolorbox}[colframe=black, colback=gray!5, boxrule=0.5mm, arc=4pt, auto outer arc]
\textit{\textbf{Q2.}} Standardization \textit{V.S.} Loss Balancing ?
\end{tcolorbox}

Feature normalization is essential in the teacher embedding space, as each teacher is trained on different datasets, causing variations in how embeddings scale and represent data. This can hinder knowledge transfer. To address this, PHI-S~\cite{Ranzinger2024PHISDB} balances the target distribution variance with the student’s estimation error variance in a label-free training setting, allowing the student model to learn effectively from diverse teacher outputs. Additionally, AM-RADIO 2.5~\citep{Ranzinger2023AMRADIOAV} shows that traditional training strategies often focus too much on a single model, leading to suboptimal performance. PHI-S standardization helps mitigate this by ensuring more balanced energies from different models.
In our approach, we combine feature standardization with our proposed loss balancing strategies, as described in Section~\ref{sec: Agglomeration Strategies}. This combination is particularly effective, as it allows for a more harmonious integration of the various teacher outputs. It is clear that our advanced agglomeration strategies can help the tokens balance better for later generalization compared to the baseline training methods. Notably, our Attention-based Loss Balancing improves performance by over 2\% in the Dice coefficient while requiring only minimal additional training parameters, demonstrating its efficiency and effectiveness in enhancing model performance.

\begin{tcolorbox}[colframe=black, colback=gray!5, boxrule=0.5mm, arc=4pt, auto outer arc]
\textit{\textbf{Q3.}} Is MedCLIP still a good pretrained encoder to learn from for the segmentation task?
\end{tcolorbox}

\begin{table}[t]
\caption{Ablation study on using different Loss Balancing strategies and feature standardization/normalization methods.}
\label{tab: loss balancing}
\begin{adjustbox}{width=230pt,center}
\begin{tabular}{c|c|c|c|c|c|c}
\toprule
\multirow{2}{*}{Row} & \multicolumn{2}{c|}{Loss Balancing} & \multicolumn{2}{c|}{Standardization}  & \multicolumn{2}{c}{VFSS-5k} \\
\cmidrule{2-7}
& \multicolumn{1}{c|}{MLP} & \multicolumn{1}{c|}{Attn} & \multicolumn{1}{c|}{$L^2$} & \multicolumn{1}{c|}{PHI-S} & \multicolumn{1}{c|}{avg. Dice $\uparrow$} & \multicolumn{1}{c}{avg. HD95 $\downarrow$} \\
\midrule
\smallcircled{1}   & \multicolumn{1}{c|}{-} & \multicolumn{1}{c|}{-} & \multicolumn{1}{c|}{\checkmark} & \multicolumn{1}{c|}{-} & \multicolumn{1}{c|}{0.800} & \multicolumn{1}{c}{14.399} \\
\smallcircled{2}   & \multicolumn{1}{c|}{\checkmark} & \multicolumn{1}{c|}{-} & \multicolumn{1}{c|}{\checkmark} & \multicolumn{1}{c|}{-} & \multicolumn{1}{c|}{0.807} & \multicolumn{1}{c}{13.654} \\
\smallcircled{3}   & \multicolumn{1}{c|}{\checkmark} & \multicolumn{1}{c|}{-} & \multicolumn{1}{c|}{-} & \multicolumn{1}{c|}{\checkmark} & \multicolumn{1}{c|}{0.811} & \multicolumn{1}{c}{13.069} \\
\smallcircled{4}   & \multicolumn{1}{c|}{-} & \multicolumn{1}{c|}{\checkmark} & \multicolumn{1}{c|}{\checkmark} & \multicolumn{1}{c|}{-} & \multicolumn{1}{c|}{0.818} & \multicolumn{1}{c}{12.947} \\
\smallcircled{5}   & \multicolumn{1}{c|}{-} & \multicolumn{1}{c|}{\checkmark}  & \multicolumn{1}{c|}{-} & \multicolumn{1}{c|}{\checkmark} & \multicolumn{1}{c|}{\colorbox{blue!10}{0.826}} & \multicolumn{1}{c}{\colorbox{blue!10}{12.641}} \\
\bottomrule
\end{tabular}
\end{adjustbox}
\vskip -10pt
\end{table}

We conducted experiments to determine the optimal number of teachers, as each imparts distinct knowledge. From \autoref{tab: teachers}, incorporating MedCLIP into the teacher pool yields only slight performance improvements. This aligns with our findings from the $LD_C$ decoder, which visualizes cosine similarity between images. In the VFSS-5k dataset, the average inter-cluster distance is 0.21, while the intra-cluster distance is 0.18. For general images, such as those in ImageNet~\cite{Russakovsky2014ImageNetLS}, cosine similarity distances range from 0.2 to 0.5~\cite{freiberger2024foolingcontrastivelanguageimagepretrained}. Thus, MedCLIP contributes minimally to the latent space with high similarity among medical image in our training data regime. Future work should explore MedCLIP's performance on larger, more diverse datasets, especially for general-purpose medical image tasks.

\begin{table}[t]
\caption{Ablation study on the selection of teacher models. The light model used is TinyViT.}
\label{tab: teachers}
\begin{adjustbox}{width=230pt,center}
\begin{tabular}{c|c|c|c|c|c}
\toprule
\multirow{2}{*}{Row} & \multicolumn{3}{c|}{Teacher} & \multicolumn{2}{c}{VFSS-5k} \\
\cmidrule{2-6}
& \multicolumn{1}{c|}{MedSAM2} & \multicolumn{1}{c|}{RAD-DINO} & \multicolumn{1}{c|}{MedCLIP} & \multicolumn{1}{c|}{avg. Dice $\uparrow$} & \multicolumn{1}{c}{avg. HD95 $\downarrow$}\\
\midrule
\smallcircled{1}   & \multicolumn{1}{c|}{\checkmark} & \multicolumn{1}{c|}{-} & \multicolumn{1}{c|}{-} & \multicolumn{1}{c|}{$0.819$}  & \multicolumn{1}{c}{13.621}\\
\smallcircled{2}   & \multicolumn{1}{c|}{\checkmark} & \multicolumn{1}{c|}{\checkmark} & \multicolumn{1}{c|}{-} & \multicolumn{1}{c|}{$0.823$} & \multicolumn{1}{c}{13.992}\\
\smallcircled{3}   & \multicolumn{1}{c|}{\checkmark} & \multicolumn{1}{c|}{\checkmark} & \multicolumn{1}{c|}{\checkmark} & \multicolumn{1}{c|}{\colorbox{blue!10}{$0.826$}} & \multicolumn{1}{c}{\colorbox{blue!10}{$12.641$}}\\
\smallcircled{4}   & \multicolumn{1}{c|}{\checkmark} & \multicolumn{1}{c|}{-}  & \multicolumn{1}{c|}{\checkmark} & \multicolumn{1}{c|}{$0.817$} & \multicolumn{1}{c}{14.007}\\
\smallcircled{5}   & \multicolumn{1}{c|}{-} & \multicolumn{1}{c|}{\checkmark} & \multicolumn{1}{c|}{\checkmark} & \multicolumn{1}{c|}{0.801} & \multicolumn{1}{c}{14.937} \\
\bottomrule
\end{tabular}
\end{adjustbox}
\end{table}
\vskip -10pt

\section{Conclusion}
\label{sec:summary}

In this paper, we present a novel approach for multi-vision model agglomeration that effectively combines knowledge from diverse teacher models into a lightweight student architecture. Our method demonstrates significant improvements in medical image segmentation tasks, achieving enhanced segmentation quality while maintaining faster inference speeds compared to individual teacher models. The proposed agglomeration strategies, particularly our attention-based loss balancing mechanism, enable efficient knowledge transfer across different visual representation paradigms. Future work will consider investigating the sequential application of teacher embeddings and the task-specific training of the efficient encoder. 

{
    \small
    \bibliographystyle{ieeenat_fullname}
    \bibliography{main}
}

\input{sec/X_suppl}

\end{document}

%% file: sec/X_suppl.tex
\clearpage
\setcounter{page}{1}
\maketitlesupplementary

\begin{figure*}[ht!]
    \centering
    \includegraphics[width=500pt]{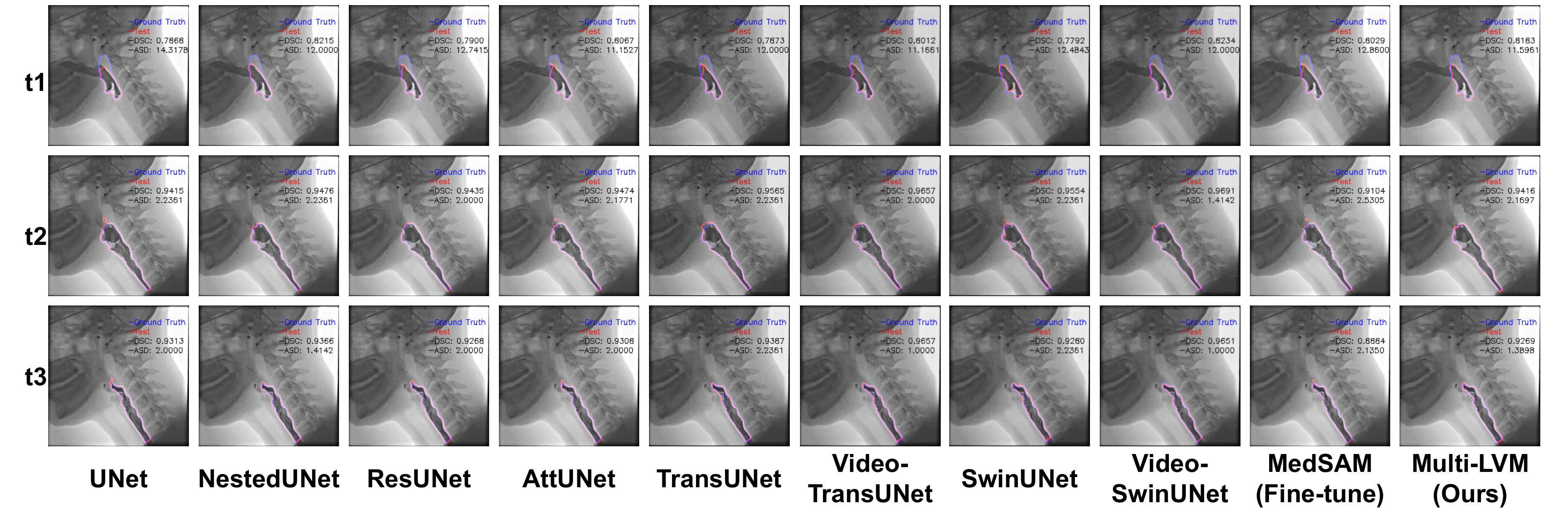}
    \caption{
Qualitative results on Pharynx Segmentation comparing Specialist Models to foundation models and our multi-encoder distilled model. The lightweight model evaluated is TinyViT.}
\label{fig:methodology}
\end{figure*}

\section{Architectural details of the encoders}
\label{Appendix: Architectural details}
The architectural details for the  MedSAM2 Encoder HieraViT-Large, and the student models, EfficientViT, TinyViT, and RepViT, are described below. The parameter size of each model are shown in the brackets. All output feature embeddings are in shape of $256\times64\times64$.

\textbf{HieraViT-Large~\cite{Ryali2023HieraAH} (224.4M).}  We follow the same architecture in SAM2. The model has channel size of $[144, 288, 576,1152]$, block size of $[2,6,36,4]$ and head size of $[2,4,8,16]$. A feature pyramid network~\cite{Lin2016FeaturePN} are used to fuse stride 16 and 32 features from Stages 3 and 4.

\textbf{EfficientViT~\cite{cai2022efficientvit}  (30.8M).} The model adopts the backbone-head/encoder-decoder design. The modules are consists of multi-scale linear attention module and an depthwise convolution. The width size is $[32, 64, 128, 256, 512]$ and depth of $1, 1, 1, 4, 4]$. The adaptor channels are $[512, 256, 128]$ and head size of $256$.

\textbf{TinyViT-Base~\cite{tiny_vit} (6.0M).} TinyViT has very similar architectural design to the original ViT and adopted ideas from SwinViT~\cite{Liu2021SwinTH}. The embedding size are  $[64, 128, 160, 320]$, with the depth of $[2, 2, 6, 2]$, and number of heads for each embedding are $[2, 4, 5, 10]$, respectively.

\textbf{RepViT-Base~\cite{Wang2023RepViTSAMTR} (5.1M).} RepViT is a pure lightweight CNN based model. The CNN kernel sizes are $3\times3$ and the channel sizes are $[48, 96, 192, 384]$ with the repetition of $[5, 6, 26, 5]$.

Training was conducted on 8 NVIDIA Tesla V100 GPUs (32 GB VRAM each) instances, with each GPU processing 4 batch size on a full resolution ($512 \times 512$) image. The inference speed was evaluated on a single NVIDIA 3090 GPU (24 GB VRAM). All code was developed in Python 3.9 and PyTorch 2.0.0 with CUDA version 12.1 and tested on Ubuntu 22.04. The code base will be released upon acceptance. Any third party codes used are licensed for research use.

\section{ACKNOWLEDGEMENTS}
 Data usage and publication are granted by UoB Ethics Approval REF: 11277. We thank project investigators David, Smithard, Ian Swaine, Salma Ayis, Aoife Stone-Ghariani, Dharinee Hansjee, Stefan T Kulnik, Peter Kyberd, Elizabeth Lloyd-Dehler, William Oliff, Lydia Morgan and Russel Walker and thank Yuri Lewyckyj and Victor Perez for their annotations. Project: CTAR-SwiFt; Funder: NIHR; Grant: PB-PG-1217-20005.